\documentclass[letterpaper, 10 pt, conference]{ieeeconf}  

\IEEEoverridecommandlockouts                              

\usepackage{graphicx} 
\usepackage{float}  
\usepackage{comment}
\usepackage{booktabs}
\usepackage{amsmath}
\usepackage{hyperref}
\usepackage{amssymb}
\usepackage{cite}
\title{\LARGE \bf
NavDreams: Towards Camera-Only RL Navigation Among Humans
}

\author{Daniel Dugas, Olov Andersson, Roland Siegwart and Jen Jen Chung
\thanks{This work was supported by the EU H2020 project CROWDBOT under grant nr. 779942.}
\thanks{The authors are with the Autonomous Systems Lab, ETH Z{\" u}rich, Z{\"u}rich 8092, Switzerland. {\tt\small\{dugasd; nandersson; rsiegwart; chungj\}@ethz.ch}}%
}

\begin{document}

\maketitle
\thispagestyle{empty}
\pagestyle{empty}

\begin{abstract}

Autonomously navigating a robot in everyday crowded spaces requires solving complex perception and planning challenges.
When using only monocular image sensor data as input, classical two-dimensional planning approaches cannot be used. %
While images present a significant challenge when it comes to perception and planning, they also allow capturing potentially important details, such as complex geometry, body movement, and other visual cues.
In order to successfully solve the navigation task from only images, algorithms must be able to model the scene and its dynamics using only this channel of information.
We investigate whether the world model concept, which has shown state-of-the-art results for modeling and learning policies in Atari games as well as promising results in 2D LiDAR-based crowd navigation, can also be applied to the camera-based navigation problem.
To this end, we create simulated environments where a robot must navigate past static and moving humans without colliding in order to reach its goal.
We find that state-of-the-art methods are able to achieve success in solving the navigation problem, and can generate dream-like predictions of future image-sequences which show consistent geometry and moving persons.
We are also able to show that policy performance in our high-fidelity sim2real simulation scenario transfers to the real world by testing the policy on a real robot.
We make our simulator, models and experiments available at \url{https://github.com/danieldugas/NavDreams}.

\end{abstract}

\begin{figure}
    \centering
    \includegraphics[width=\linewidth]{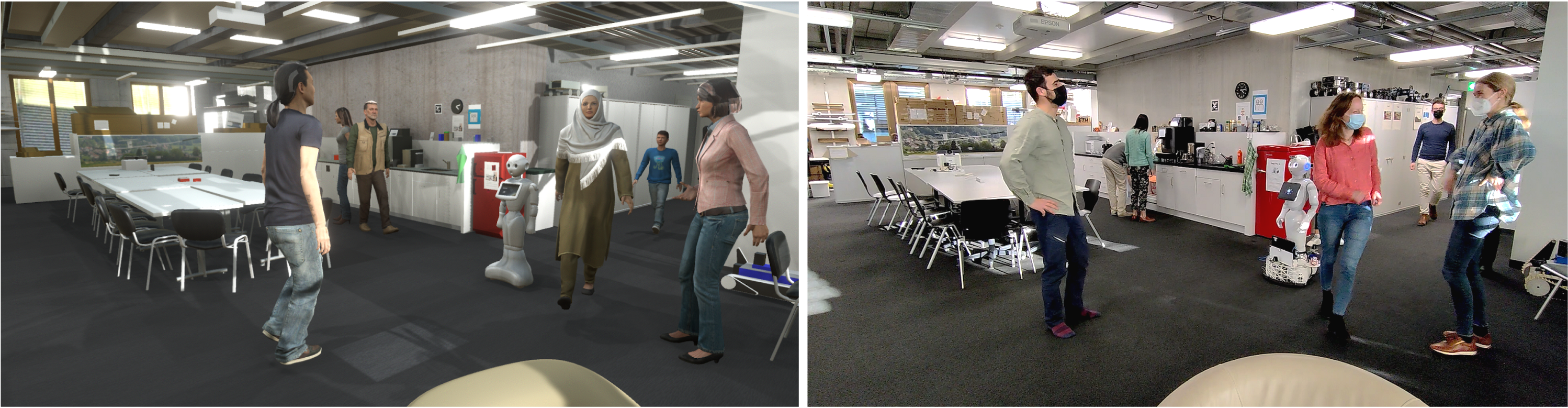}
	\caption{Testing our proposed camera-based RL policy in (left) simulation and (right) real world. We find that our high-fidelity simulator allows sim2real transfer.}
	\label{fig:title_comparison}
\end{figure}

\section{Introduction}


The ability to safely move in uncontrolled environments such as public spaces or offices is a key requirement of autonomous robots. As a result it is subject to intense interest from both industry and academia. Due to their unstructured nature, these spaces pose many challenges which make planning difficult. One key challenge is the presence of humans, who display complex and dynamic behaviors. 

Though navigation planning is typically done using distance sensors such as LiDAR or depth, due to the ubiquity of monocular cameras and their low price, there is increasing research interest in solving similar problems using images, despite the increased difficulty that this incurs.
One interesting property of image-based navigation is that it is typically higher dimensionality than other low-cost options such as 2D LiDARs or proximity sensors, having the potential to capture more information from the scene, such as body motion cues.
%

Since crowd navigation is fundamentally about selecting the best action and reinforcement learning (RL) has shown success on other vision-based planning tasks \cite{mnih2015human}, using RL for crowd navigation from monocular camera is an avenue of research with considerable potential. In this paper we make multiple contributions to aid research in this area, with the goal of serving as a stepping stone towards realizing efficient RL-based solutions to the crowd navigation problem using only monocular camera. These contributions range from a proposed simulator (Fig.~\ref{fig:title_comparison}), to exploring alternatives to standard end-to-end baselines, as well as sim2real transfer capabilities. We further explore the use of the recently proposed \emph{world-models}~\cite{ha2018recurrent} architecture on this problem, and show that it learns useful representations for navigation and improves performance over the baseline end-to-end approach. 


\begin{figure}
    \centering
    \includegraphics[width=\linewidth]{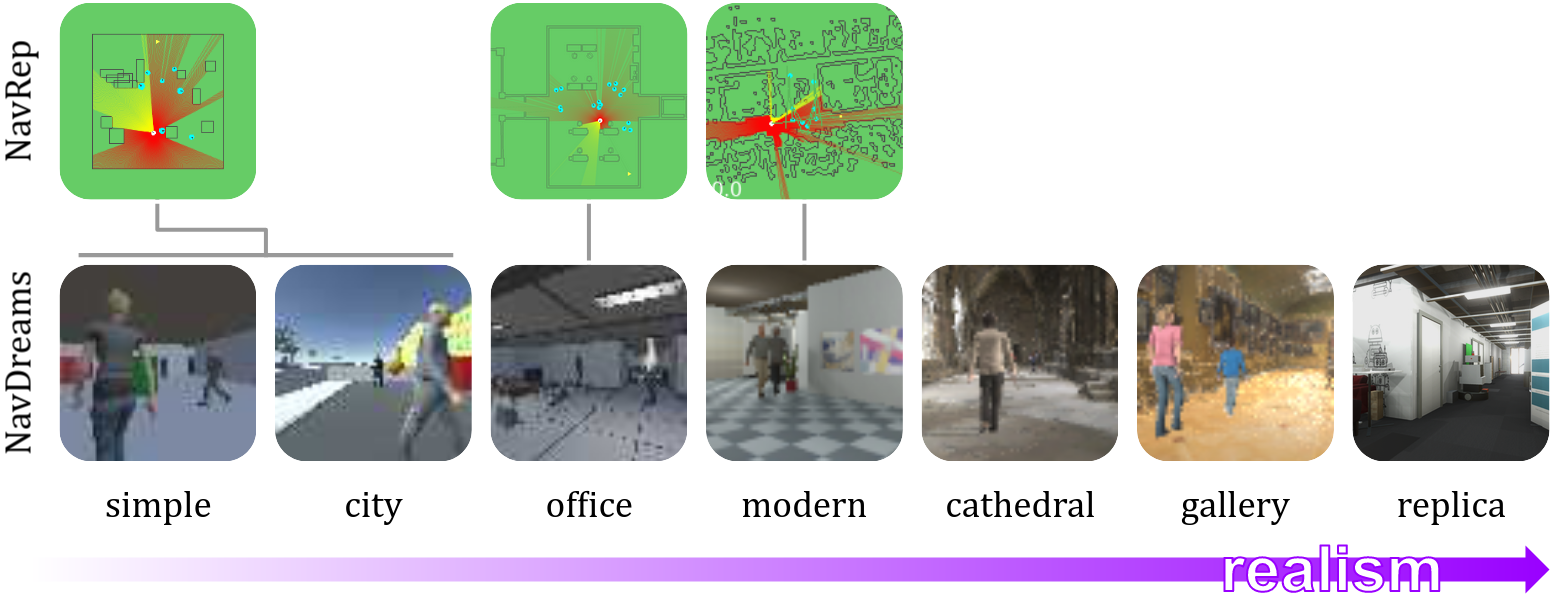}
	\caption{Available environments in the NavDreams simulator, and the 2D navigation environments they are based on.}
	\label{fig:envs2}
\end{figure}

Our contributions are:
\begin{enumerate}
    \item We design an open-source simulator for camera-based robot navigation among humans. Included are several environments ranging from classic path planning baselines to realistic human environments (Fig.~\ref{fig:envs2}).
    \item We show that world-model-based RL approaches are able to navigate successfully around humans using only monocular vision.
    \item We show that state-of-the-art world-model approaches can be used to generate dream-like navigation sequences and encode features useful for planning.
    \item We apply sim-to-real methods and show that the learned policies are able to transfer successfully from simulation to a real robot (Fig.~\ref{fig:title_comparison}).
\end{enumerate}

\section{Related Work}

Recent works in deep reinforcement learning have proposed solutions for image-based navigation, producing policies that can traverse simulated and real-world environments. Typically these methods exploit additional structural elements such as auxiliary task learning~\cite{mirowski2017learning} or explicitly incorporate the idea of landmarks into the navigation learning architecture~\cite{savinov2018semi} to improve performance. In terms of applying these methods to real-world environments, \cite{mirowski2018learning} leveraged the vast amount of Google StreetView data to enable end-to-end learning for mapless navigation (albeit still in the StreetView environment and not in reality), while \cite{zhu2017target} demonstrated that fine-tuning policies trained in high fidelity simulations resulted in the best performance transfer to a real robot.

A related class of methods has also focused on learning camera-based exploration of unknown environments~\cite{ramakrishnan2021exploration}. For example, \cite{chen2019learning} proposed an end-to-end approach that applies map coverage as the learning signal. As an alternative to end-to-end learning, \cite{chaplot2020learning} combined learned modules within a classical navigation pipeline to produce the 2019 winning entry in the Habitat PointGoal navigation challenge \cite{savva2019habitat,kadian2020sim2real}.

Nevertheless, each of the aforementioned methods only targets navigation in static environments. Dynamic environments, specifically those with human pedestrians, introduce an additional layer of complexity for learning navigation strategies. Collision avoidance becomes more challenging, all the more so since the interactive and social nature of human motion can be especially difficult to predict. Due to this, most existing works opt to focus on the pure collision avoidance aspects which can be adequately captured via lower-dimensional range-based sensing inputs (e.g. LiDAR)~\cite{chen2017decentralized, chen2017socially, everett2018motion, chen2019crowd, liu2020soadrl, fan2020distributed, dugas2021navrep}. Furthermore, works such as~\cite{chen2017decentralized, chen2017socially, everett2018motion, chen2019crowd, liu2020soadrl}, which include aspects of social compliance, assume that positions of nearby agents are known. In contrast, our prior work~\cite{dugas2021navrep} showed that world-model-based RL could provide the representational capacity needed to accurately predict, or ``dream'', future LiDAR observations and use these representations to learn effective autonomous crowd navigation policies without the need for explicit human pose information.

World-model-based RL separates the prediction (the world-model itself) and control elements of RL~\cite{ha2018recurrent}. As such, the world-model can be pre-trained offline, allowing a lighter architecture for the controller, which is trained online. An additional benefit is that once the world-model is trained to produce accurate dreams, the dreams can be directly used to train the controller, supplementing or potentially altogether bypassing the need for expensive data collection in simulation or the real world. Apart from our use in LiDAR-based navigation, this concept has also been successfully applied to learn policies with state-of-the-art performance on Atari benchmarks~\cite{hafner2021dreamerv2} as well as for camera-based robot navigation in static environments~\cite{piergiovanni2019learning}. These results were especially promising and motivated our present work to investigate the potential of world-model RL with camera-only inputs for learning the complex task of navigating among dynamic human crowds.

\begin{figure}
    \centering
    \includegraphics[width=\linewidth]{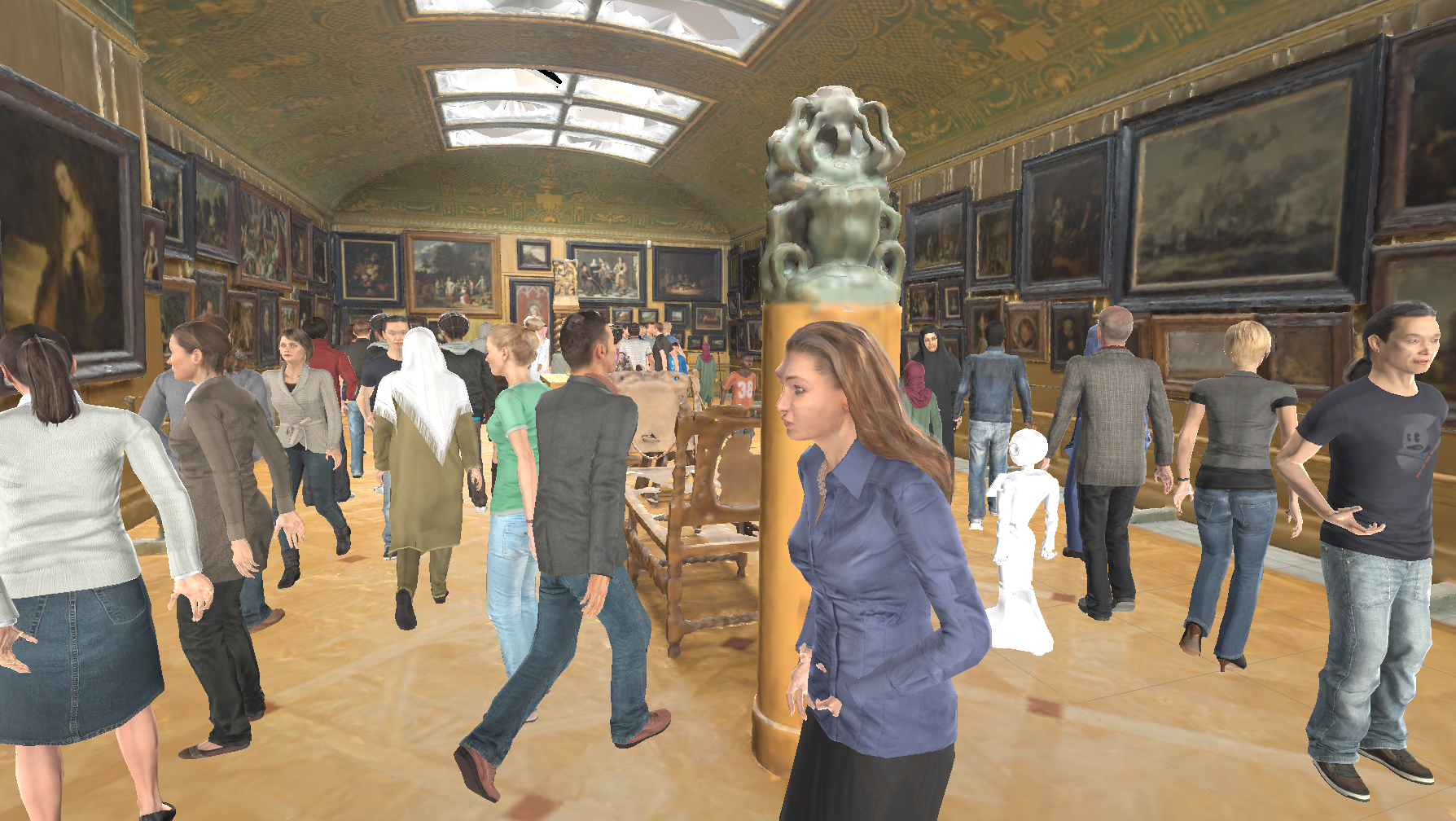}
	\caption{Example crowded scene in the \textit{gallery} environment with 50 simulated pedestrians.}
	\label{fig:crowded}
\end{figure}  

\begin{figure*}
    \centering
    \includegraphics[width=\linewidth]{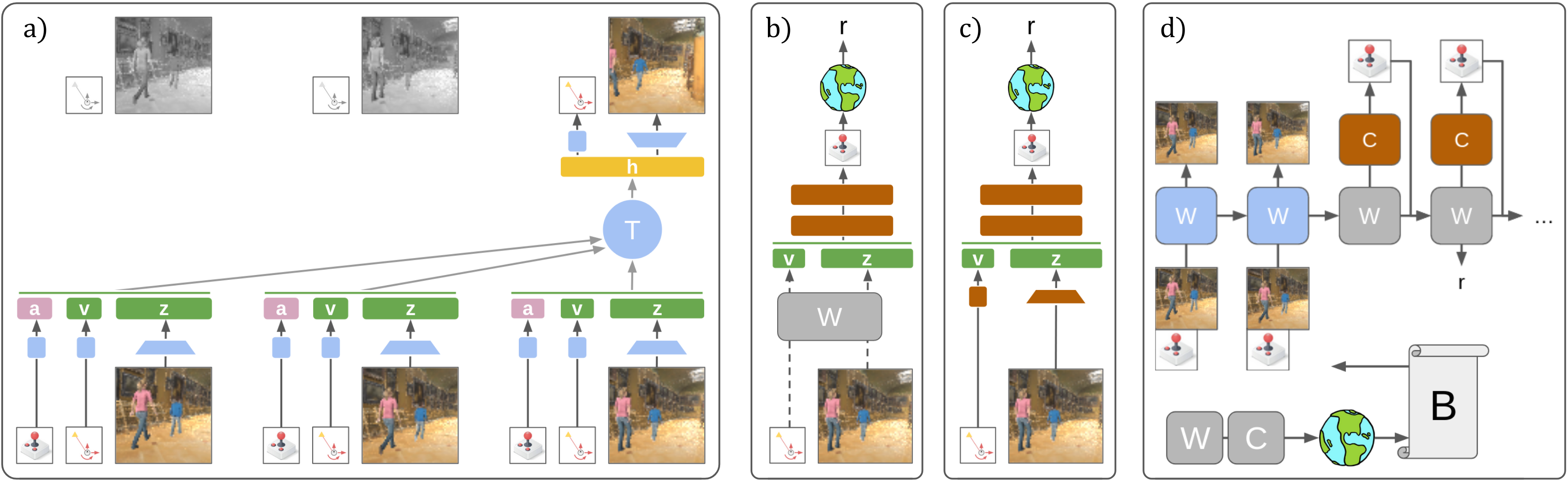}
	\caption{ Illustrations of a) the Transformer-based world-model training process. Sequences of images, actions (joysticks) and vector observations are encoded into their respective features (z, a, v), which the causal self-attention blocks (T) use to emit a predicted feature encoding which is decoded into an image and vector observation prediction for the next time-step. b) The single-step controller training process for the controller using world-model features and c) for the end-to-end controller d) illustrates the more complex explicit model-based training process used when testing the Dreamer approach. In contrast to b) and c) where the policy actions are applied to the environment (planet symbol) to obtain rewards (r) used as training signals, in d) rollouts are collected in the environment and stored in a database (B). This database is used to train the world-model, which then generates dream sequences. These sequences, which include rewards, are in turn used to train the controller. The W (world-model) and C (controller) modules are greyed out to denote frozen weights.}
	\label{fig:world_model}
\end{figure*}

\section{NavDreams Simulation Environment}

To learn camera-based navigation policies, we require access to a representative training environment. However, it is impractical (and unsafe) to generate robot-crowd navigation experiences directly in the real world.
Furthermore, existing 3D simulators do not provide models of multiple moving pedestrians~\cite{xia2018gibson, chang2017matterport3d, zhu2017target, tolani2021visual}, whereas those that do provide human models are focused on general high-level planning and manipulation, and do not offer models or scenarios tailored to robot navigation research \cite{puig2018virtualhome}.
%
Thus, we create, and make available, the NavDreams simulation environment which aims to reproduce typical crowd navigation scenarios.

The simulator allows varying the difficulty level of the navigation task within a single scenario. This difficulty is represented as an arbitrary scalar. The higher the difficulty, (i) the further away goals are sampled from the robot, (ii) the more humans are spawned into the scene, and (iii) the more static obstacles are present.
We also provide increasing levels of realism such as
camera shake,
image post-processing,
robot inertia,
increased scene complexity and
multiple people behaviors.
Our simulator is provided as open-source code, and its parameters (camera height, base velocity, camera shake intensity, FOV) can be varied to reproduce most robot configurations.
It works out-of-the-box with the widespread RL OpenAI gym API.
It can also be used to generate depth and segmentation labels for the simulated camera images. Example scenarios can be seen in Fig.~\ref{fig:title_comparison} and Fig.~\ref{fig:crowded}.

\subsection{Scenarios}

The full set of simulator scenarios used in this paper are shown in Fig.~\ref{fig:envs2}. They are:

\subsubsection{simple}
A 3D re-interpretation of the classic 2D scenario found in \cite{dugas2021navrep,chen2019crowd}. It includes procedural obstacle generation and spawning of moving humans, as well as random initial robot orientation.

\subsubsection{city}

Identical to the simple scenario except that the walls and background are replaced with a static city scene.

\subsubsection{office}

A reconstruction of the office scene used in \cite{dugas2021navrep,dugas2020ian}. Unlike in \textit{simple} and \textit{city}, the initial positions of the robot and humans are predetermined for each difficulty level, and are not procedurally generated.
In addition to walls, this scenario contains commonplace objects, such as tables, chairs and plants, which act as static obstacles.

\subsubsection{modern} 

An indoor office scene with a museum-like section.
Robot, human spawns and goals are randomly sampled.
It is significantly more complex than the previous three scenarios due to having a figure eight layout, much larger variety of common objects, higher maximal crowd density and more complex human behaviors (randomly switching between talking, walking, idling). We also include real-world robot effects such as inertia in the robot movement and camera shake caused by changes in velocity.
\subsubsection{cathedral}

A reproduction of the Cloister Cathedral of St. Mary of La Seu Vella, from a publicly available 3D model captured through photogrammetry, using a Sony a7R II camera~\cite{calidos2020cloister}.
The crowd simulation is identical to \textit{modern}.

\subsubsection{gallery}

A reproduction of the gallery room of the Hallwyl museum, also captured through photogrammetry (Panasonic Lumix GX8 camera)~\cite{lernestal2019museum}.
Though the crowd simulation is identical, being smaller than the \textit{cathedral} scenario, higher crowd densities are reached at lower difficulty levels as shown in Fig.~\ref{fig:crowded}.

\subsubsection{replica}

A hand-made model of our office common space, reproduced with as much fidelity as possible for sim2real experiments. Some of the scene objects are randomly moved at the start of each episode. Lighting is randomly sampled from a set of light conditions (indoor and outdoor light sources turned off-on, color shifts). Post-processing is added for photorealism and exposure is approximately tuned to match typical robotics camera sensors (e.g. Intel RealSense d415).
A comparison between the simulation and reality is shown in Fig.~\ref{fig:title_comparison}.


\section{Learning Camera-based Navigation}

\subsection{Problem Definition}


\subsubsection{State space}
The state available to the model consists of the \textbf{image observation}, a single RGB 64x64 image from the robot's forward facing camera and a \textbf{vector observation}, a two-element vector containing the goal's coordinates in meters in the robot's frame.

\subsubsection{Action space}
In preliminary testing with continuous actions and image inputs, baseline end-to-end approaches struggled to converge.
As a result, unlike in \cite{dugas2021navrep}, we opt for a discrete action space of size 3: [LEFT, FORWARD, RIGHT].
LEFT and RIGHT correspond to rotations around the vertical axis.

\subsubsection{Reward function}

The stepwise reward is the sum of four terms:
$r_g$ is 100 if the goal is reached, 0 otherwise;
$r_c$ is -25 if the robot is in collision, 0 otherwise;
$r_p$ is the straight-line progress towards the goal in meters since the last step multiplied by 0.1;
$r_t$ is -0.05 if the action is LEFT or RIGHT to penalize superfluous turning.

In \textit{modern, cathedral, gallery} and \textit{replica} scenarios, a simpler reward function is used, where
$r = r_g$. We found that this was a sufficient reward signal for learning successful crowd navigation policies.

\subsection{World-models}


Like in \cite{dugas2021navrep} the full architecture is split into several modules. Here we consider the world-model module \textbf{W}.

\subsubsection{Architecture}
The world-model takes sequences of images, vector observations, and continuous actions (3-vector, forward, sideways, and rotational velocity). It learns a mapping from this input sequence to latent features $z$, $h$, as shown in Fig.~\ref{fig:world_model}a, and from these latent features to a next-image and next-vector observation prediction. Taking in continuous actions (rather than just the discrete actions used by the RL controller) allows us to use any available robot data that includes robot base velocities and camera images (e.g. ROSbags from joystick-operated runs) to train the world-model offline. The [LEFT, FORWARD, RIGHT] motions are simply converted to continuous velocity commands when passed through the world-model.


\textbf{W} is implemented using fully connected layers to encode the action and vector observation, a CNN to encode image inputs and a transformer T to compute a prediction from the sequence. A latent feature size of 64 is used for the image encodings $z$ as well as for the prediction encodings $h$.

\subsubsection{Loss}

The loss used to train the transformer-based \textbf{W} module is, \begin{equation}\label{eq:pred_loss}
    L_{W} = L_{\textit{prediction}} + L_{\textit{reconstruction}} + \kappa KL(\hat{z}).
\end{equation}

The prediction $L_{\textit{prediction}}$ and reconstruction $L_\textit{reconstruction}$ losses are defined as the mean-squared-error between target and predicted image.
The $KL(\hat{z})$ term is the Kullblack-Leibler divergence between
the latent feature distribution parameterized by the encoder
and the prior latent feature distribution, which we multiply by a constant coefficient. In practice we use a small value (0.001) for $\kappa$.

\subsubsection{Datasets used}\label{sec:dataset_variants}
In order to train the world-model module \textbf{W}, we use both real-world data and data generated with the simulator. In the real-world dataset, the robot is either driven manually or by a conventional planner. In the generated data, we use a simple stochastic heuristic policy, which on average drives towards the goal, but may deviate from these actions at random (noise on the velocity commands drawn from a Gaussian distribution).

Rather than train a single world-model on all available data, we split the data into subsets and train several world-model variants. This allows us to examine the impact of dataset scope on final performance.
These variants are referred to as S, SC, SCR, and R, based on the scenario data they were trained with.
S was trained only on data from the \textit{simple} scenario.
SC from the \textit{simple}, \textit{city} and \textit{office} scenarios.
R on data from real recordings plus data from the \textit{modern} scenario.
SCR was trained on all of the above scenarios.

\subsubsection{Architecture variants}\label{sec:architecture_variants}

In addition to the transformer world-model architecture mentioned above, we experiment with other SOTA world-model architectures:
A Recurrent Stochastic State Model (RSSM) architecture, as proposed in \cite{hafner2019learning} and used in \cite{hafner2021dreamerv2}.
A Transformer Stochastic State Model (TSSM) as proposed by \cite{chen2022transdreamer} is also evaluated in this work.

The RSSM and TSSM are trained using the same loss functions as was used in \cite{hafner2021dreamerv2}, where the loss is the sum of the reconstruction loss, and the KL divergence between posterior and prior stochastic state distributions.

The same layer and latent feature dimensions were used as in the original works. When not otherwise stated, we use the same transformer latent feature, embedding, attention heads and layer dimensions as in \cite{dugas2021navrep}. We also test a transformer world-model with a larger embedding and latent feature size of 1024 to match that of the RSSM, and TSSM.

\subsection{Controller}

The controller module \textbf{C}, takes as input the world-model latent feature vector for a single timestep, and outputs a single action for that timestep. It is implemented as an MLP with two fully connected layers of size 1024. 
All policies take the given goal vector $v$ as input in addition to the different world encodings below.

\subsubsection{Pre-trained world-model (WM+PPO)}

For most experiments, we use the straightforward controller training method from \cite{dugas2021navrep}, in which the most recent state $z_t$ of a pre-trained world model \textbf{W} is used as input to the policy, which is trained in simulation with PPO \cite{schulman2017proximal}.  

\subsubsection{End-to-end controller baseline (End-to-end)}

We also implement a baseline controller whose architecture consists of a feature encoder followed by an MLP, which are trained end-to-end (shown in Fig.~\ref{fig:world_model}c). To allow for fair comparison, the encoder is identical to \textbf{W}, while the MLP is identical to the one used in \textbf{C}. The end-to-end controller is also trained from single timestep observations.
%

\subsubsection{Explicitly model-based learning (WM+Dreamer)}

We also test the explicitly model-based training method proposed by \cite{hafner2021dreamerv2}.
In this method, an actor-critic policy is trained using world-model dream prediction sequences. In order to collect a rollout dataset, this policy is iteratively applied to the environment, which is in turn used to train the world-model (illustrated in Fig~\ref{fig:world_model}d).
As a consequence, the world-model \textbf{W} is not exclusively trained on a previously collected offline dataset (as is the case with the single-step controller experiments), but rather trained on-policy.

When using this method, we stay as close as possible to the original work, using the same architecture, hyperparameters and optimization procedure.

\section{Experiments \& Results}
We first analyze the single-step controller training methods, comparing the end-to-end learning baseline to using the world-model features which are trained offline. Then, in Sections~\ref{sec:world_model_architecture_compare} and \ref{sec:variations_in_controller} we compare the performance of the different world-model architectures (transformer-based vs. Dreamer). Finally, in Section~\ref{sec:sim2real} we show sim-to-real policy transfer performance on a real robot.

\subsection{Comparison with End-to-end Baseline}

We train the end-to-end controller models on all simulation environments for 5 million steps, at which point we observe that training stops improving.
The training was done with PPO \cite{schulman2017proximal} which we found to perform better on this problem than competing approaches in preliminary testing. 
%
We used a curriculum approach to adapt the environment difficulty to the model performance, increasing it by 1 increment after each successful episode, and decreasing it by 1 after each failed episode. An increase in difficulty of 1 increment involves adding one additional pedestrian and increasing the maximum initial distance to goal by 1m. 
Training is repeated with 3 random seeds, to confirm statistical relevance of the learning results.

We apply the same training process to learn a controller using the transformer-based world-model features. The world-model was trained off-policy on a dataset collected from the SCR environments as described in Section~\ref{sec:dataset_variants}.

\begin{figure}
    \centering
    \includegraphics[width=\linewidth]{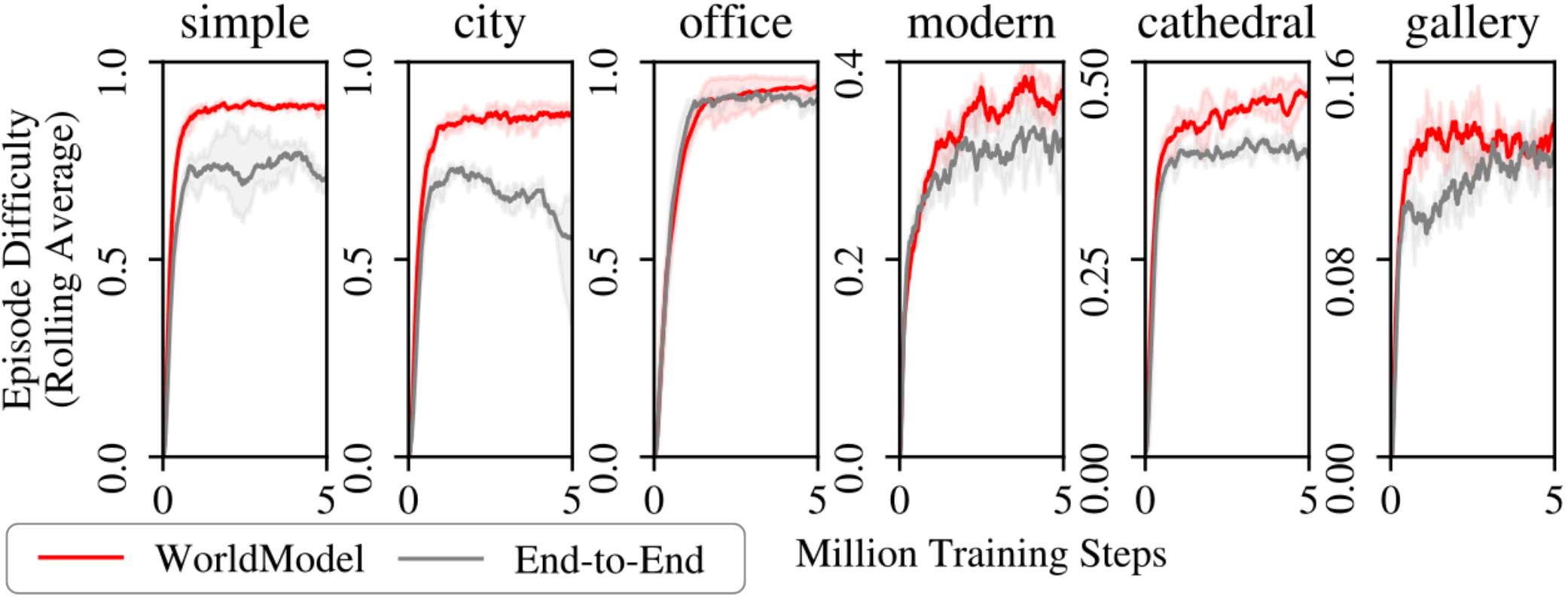}
	\caption{Comparison of end-to-end and world-model controller training performances in each simulated environment. Shaded area around each curve shows the min-max spread between seeds. In this comparison, all controllers use the same world-model, trained on SCR.} 
	\label{fig:best_of_each}
\end{figure}

\begin{figure}
    \centering
    \includegraphics[width=\linewidth]{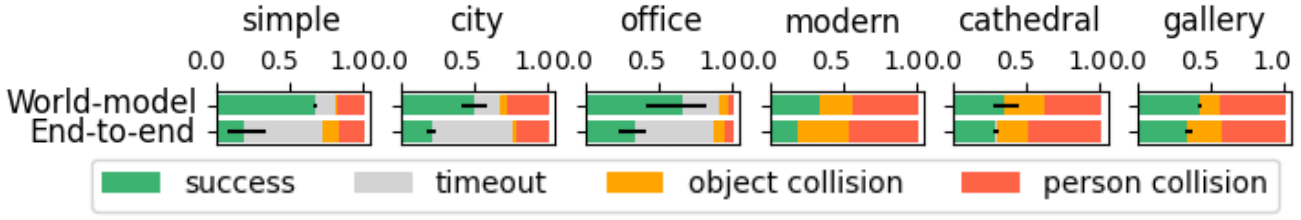}
	\caption{Test-time performance of End-to-end and World-model policies in each scenario. Min-max performance across seeds shown in black. In this comparison, all controllers use the same world-model, trained on SCR.}
	\label{fig:best_test}
\end{figure}

Fig.~\ref{fig:best_of_each} shows that the world-model approach is consistently able to reach a higher average difficulty setting during training. We also test the trained models on a fixed difficulty level
(the highest level reached by either model during training)
and find the that controller trained using world-model features outperforms the end-to-end baseline in every scenario (see Fig.~\ref{fig:best_test}).
%


\begin{figure*}
    \centering
    \includegraphics[width=\linewidth]{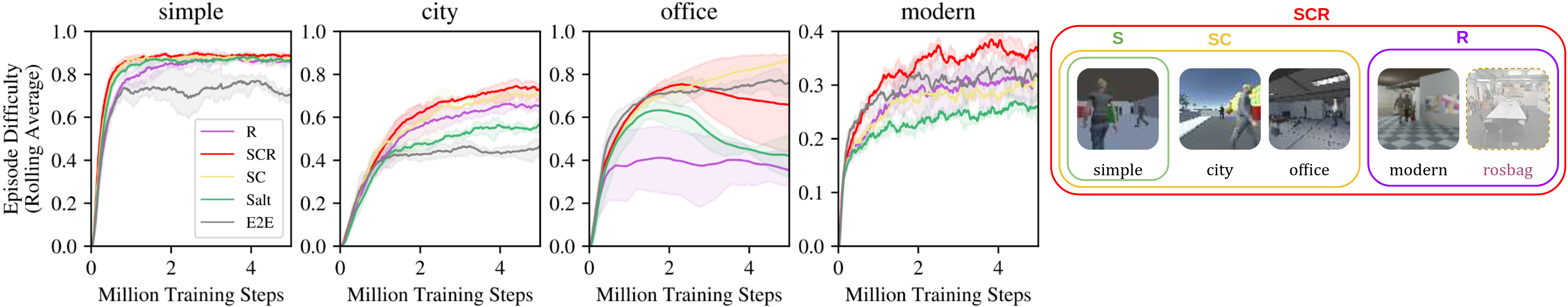}
	\caption{World-models trained with different datasets show varied training performance. The scenario dataset scopes corresponding to each W variant (S, SC, SCR, R) are shown on the right.}
	\label{fig:simple_xtrain}
\end{figure*}








\subsection{Impact of World-model Dataset}

To quantify the impact of dataset scope, we train each variant in and out of their respective evaluation scenarios.
More precisely, for each \textbf{W} variant (S, SC, SCR, R) we also train several separate \textbf{C} controllers, one on each target scenario (\textit{simple, city, office, modern}).

The results given in Fig.~\ref{fig:simple_xtrain} show two unexpected results: 1) Variants can outperform the end-to-end baseline when trained on scenarios that the world-model has never seen, and 2) we find that more general variants (\textbf{W} trained on a larger subset of data) outperform specific variants (\textbf{W} trained only on the subset relevant to the target scenario).
The performance of world-models trained with less data (S, R) can generally be improved by supplying more data even if that data does not reflect the target environment (SC, SCR).


\subsection{Impact of Controller Scope}\label{sec:controller_generalization}

\begin{figure}[t]
    \centering
    \includegraphics[width=\linewidth]{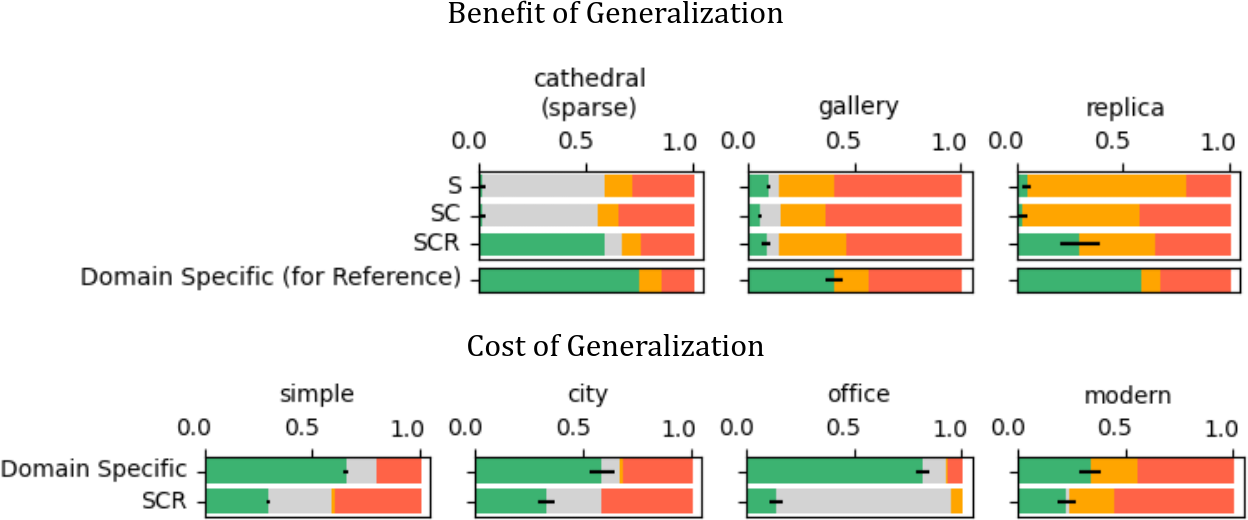}
	\caption{When tested outside of their training environments, controller performance is greatly reduced. We see that training `generalist' controllers (trained on S, SC, SCR scenario subsets) can improve generalization ability, at the cost of reduced performance when compared to scenario-specific controllers.}
	\label{fig:ood}
\end{figure}


%
To test the impact of controller scope on generalization ability, we train controllers on a mix of several scenarios, corresponding to the S, SC, and SCR scenario subsets.
Fig.~\ref{fig:ood} shows both benefit and cost when training generalist controllers:
i) Generalist controllers (e.g. trained on SCR) perform better on average than less or non-generalist controllers in unseen environments (top row).
ii) Generalisation comes at a price, generalist controllers are not able to perform as well in seen environments as their specialist counterparts, even after training converges (bottom row).

These results indicate that overfitting is a larger issue for monocular-based navigation than for LiDAR-based navigation.
This motivates the need for varied scenarios in deep-learning research for image-based navigation.

\subsection{Multi-task Experiment}

One hypothesis for the better performance of the world-model feature controller models is that the self-supervised learning leads to `richer' latent features, which can be leveraged by the controller.
By `richer' we mean that the prediction task leads to a more efficient compression of the world state information into the latent features.
%

In order to test this hypothesis, we ask whether the latent features of the pre-trained \textbf{W} module can be used for other tasks.
We pick the tasks of pixel-wise semantic segmentation and depth-prediction, in order to quantify how much semantic and geometric understanding is distilled into latent features by the model.

To perform this experiment, we first generate a dataset of semantic and depth ground truth in the simulator.
Then, we train three versions of the same decoder architecture to predict the semantic and depth labels, one taking as input the latent features from the world-model, the second the latent features from the same layer of the end-to-end baseline network. To serve as a best-case, we also train the third decoder jointly with an encoder (identical to the end-to-end and world-model encoder modules) to predict the labels from images directly.
In all cases the latent feature size and decoder architecture is identical.
%

For the segmentation task, the prediction error is calculated as the mean per-pixel binary cross entropy error between the predicted classes and the ground truth.
For the depth task, the prediction error is the per-pixel mean square of the proportional error, where the proportional error is the difference between predicted and ground truth distance divided by the ground truth distance.

\begin{figure}
    \centering
    \includegraphics[width=\linewidth]{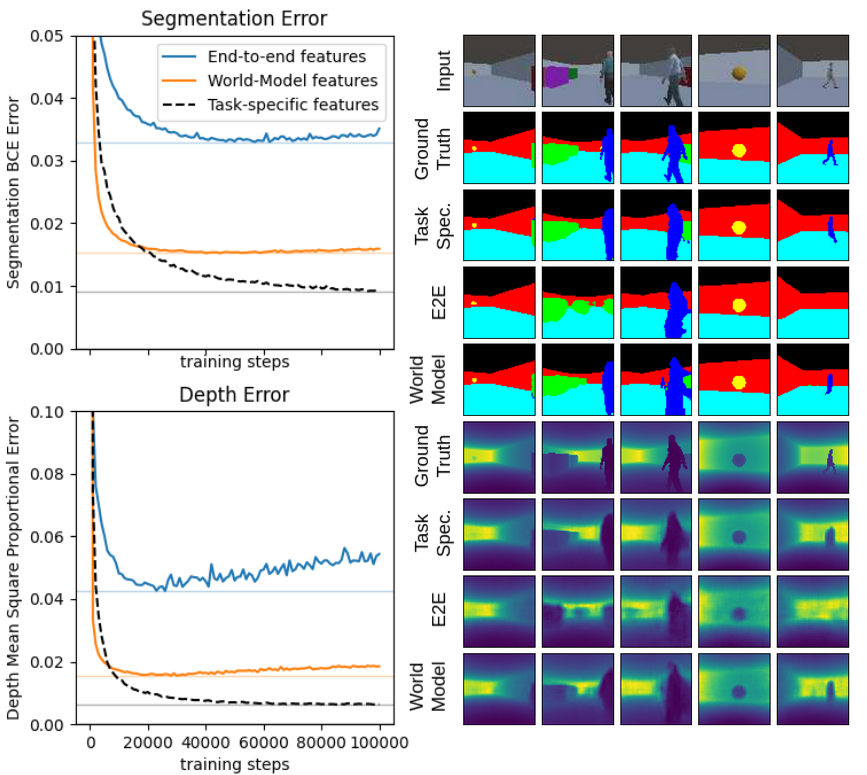}
	\caption{Quantitative (Right) and qualitative (Left) comparison of world-model and end-to-end features for the class segmentation and depth prediction tasks in the multitask experiment. Note that the E2E features tend to `lose' pedestrians.}
	\label{fig:multitask_results}
\end{figure}

On the test set for both tasks, the decoder based on world-model latent features achieves a per-pixel average error which is half that of the end-to-end latent features decoder. These results, shown in Fig.~\ref{fig:multitask_results},
%
support the latent-feature `richness' hypothesis above.









\subsection{World-model Architectures}\label{sec:world_model_architecture_compare}

To find out which world-model architectures are better suited to the navigation problem, we compare the quality of their predictions, and their impact on controller performance.

\subsubsection{Dream quality}

One of the assumptions in the world-model approach is that a model which is better able to predict future states or sensor data can better capture the world dynamics, and that this ability is inherently useful for planning, which the \textbf{C} module is meant to put into practice.
We experiment with the three architectures for the world-model described in Section~\ref{sec:architecture_variants}. The transformer architecture used in \cite{dugas2021navrep}, the RSSM architecture used by \cite{hafner2021dreamerv2}, and the TSSM architecture from \cite{chen2022transdreamer}.

\begin{figure*}
    \centering
    \includegraphics[trim=0 0 125 0, clip, width=\textwidth]{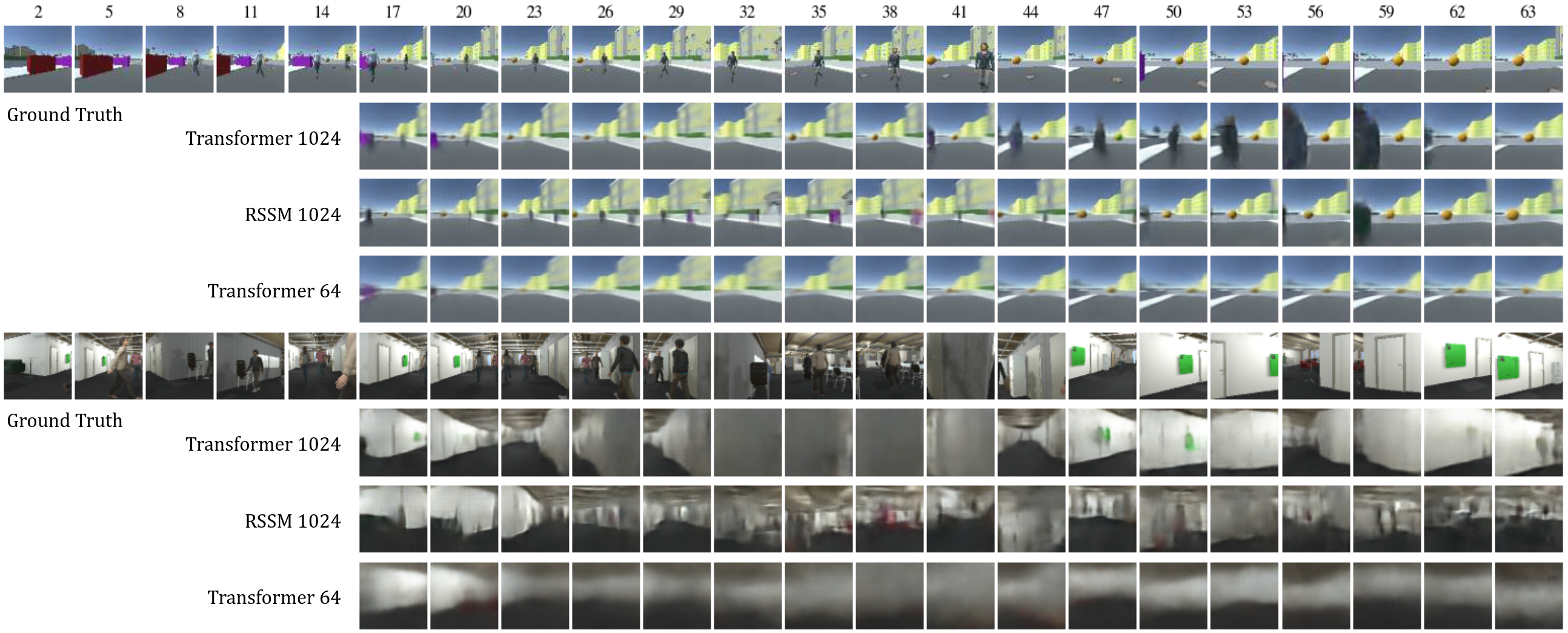}
	\caption{Predicted dream-sequences for three of the tested world-model architectures, given the same action inputs, and context. We see that the world-model approach successfully captures the appearance and dynamics necessary for navigation. Transformer and RSSM architectures with z=1024 output dreams of similar apparent quality, tracking the goal well, and with moving shadows resembling people. The small transformer creates blurrier predictions, and has a stronger tendency to remove people from its predictions.}
	\label{fig:dream_examples}
\end{figure*}

As shown in Fig.~\ref{fig:dream_examples}, the trained world-models are able to generate dreams which appear to capture dynamics  which matter for navigation (robot and people movements).

To measure both immediate and longer-term prediction ability, we calculate n-step prediction error, which compares the dream image predictions to the real sequence.
What we refer to as `dreams' are sequences of predicted observations where the latest prediction is used as input for the next, and so-on.
As such, the 1-step prediction error is the immediate prediction from the model given the input sequence, the 2-step prediction error is the prediction from that sequence with the 1-step prediction appended to it, and so on. Actions are taken from the ground truth sequence.

To give an upper bound on the errors, we calculate i) the prediction error if a model outputs constant grey images (all values 0.5), ii) the prediction error if a model just outputs the last image in the input sequence (static assumption).
This ensures that models which perform better than the upper-bound are doing better than random guessing, or predicting stillness.
The same SCR dataset is used to train all architectures, with a subset withheld for validation.
%

Qualitative results in Fig.~\ref{fig:dream_examples} show that the RSSM and 64-feature transformer better capture general dream dynamics (performance over several frames), but the large transformer creates higher-fidelity dreams (more coherent frame-to-frame, higher resolution)\footnote{TSSM produced similar quality dreams compared to RSSM and so we do not include their qualitative results here due to space restrictions but include them in the associated video.}.

\begin{figure}
    \centering
    \includegraphics[width=\linewidth]{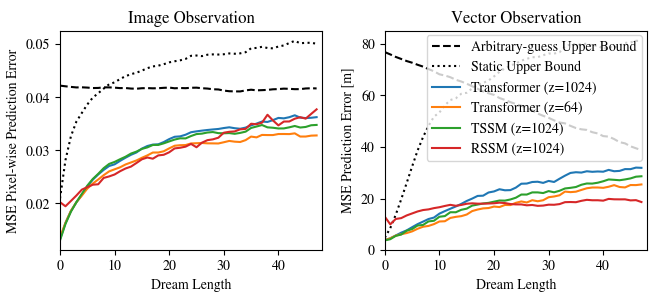}
	\caption{Mean test-set error between dreamed world-model trajectories and ground truth is computed for various world-model architectures. On the left for trajectory images, and on the right for vector observations (goal position in robot frame).}
	\label{fig:dream_error}
\end{figure}

However, the quantitative results in Fig.~\ref{fig:dream_error} do not show a similar trend; the pixel-wise error lead to lowest error for the blurriest predictions (Transformer $z=64$).
%
This shows that proper evaluation of long-term prediction quality is not straightforward, though simple metrics are able to capture the general ability of a model, there are aspects we might consider more important than pure prediction accuracy (e.g. coherence, object permanence, etc.).
%



\subsubsection{Impact on controller performance}
We compared the impact of using RSSM, TSSM, and our standard transformer world-models for training the controller and found no significant difference in performance (similar mean and variance). We speculate that these architectures are all general enough that hyperparameter search, training scaling effort and engineering effort likely play a bigger role in final performance than network architecture for this problem.


\subsection{Variations in Controller Training}\label{sec:variations_in_controller}

\subsubsection{Sequence Input to Controller}
Several SOTA end-to-end RL approaches use sequences of images as input to the controller model \cite{chen2022transdreamer}. In our experiments, using sequences (length 10) of encoded feature vectors instead of only the latest $z$ vector as controller input led to reduced performance in all simulated scenarios.
%

\subsubsection{Explicit Model-based Training}

We test the explicitly model-based learning of world-models from \cite{hafner2021dreamerv2}, incorporating the larger architectures, and on-policy world-model training. Model-based approaches are known to be more data efficient, but also very compute heavy since the policy is optimized from roll-outs of the world-model, which in turn has to be continuously updated.
Due to the increased computational cost, these models are trained to convergence only on the \textit{simple}, \textit{modern}, and \textit{replica} environments, which takes more than 1 week per model on a 16GB Tesla T4 GPU. Results are shown in Table~\ref{tab:successrates}.

\begin{table}[b]
\caption{Success rates}\label{tab:successrates}
\centering
\begin{tabular}{ c c c c c c } 
\toprule
Scenario   & Simple & Modern & Replica & Replica & Replica \\
\# people & 10 & 25 & 0 & 3 & 10 \\            
\midrule
WM+PPO &      57 \%        &       30 \%         &      94 \%   & 55 \%       &     17 \% \\ 
Dreamer (WM) &            82 \%      &         50 \%          &     96 \%    & 65 \%  &         17 \%\\ 
\bottomrule
\end{tabular}
\end{table}


We find that sizeable performance gains can be obtained from incorporating methods used in latest model-based RL approaches. 
%
These results indicate that there is likely still significant performance gain to be achieved on this problem through scaling and model tuning.
This makes us hopeful that we can bring navigation models even closer to reality and ultimately achieve human-level planning performance.

\begin{figure}
    \centering
    \includegraphics[width=0.8\linewidth]{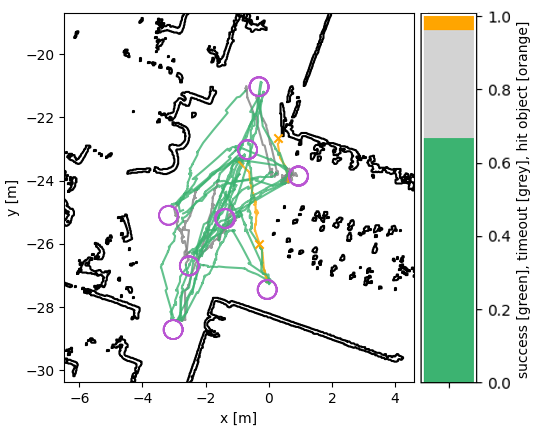}
	\caption{Robot trajectories during real-world tests (people are not shown).}
	\label{fig:irl_test_traj}
\end{figure}

\subsection{Sim-to-real}\label{sec:sim2real}
Since the controller generalization experiments in Section~\ref{sec:controller_generalization} demonstrated the benefit of training controllers on the target environment, here we use the proposed NavDreams simulator to create the \textit{replica} simulation environment shown in Fig.~\ref{fig:title_comparison}, reproducing the chosen real test conditions as realistically as possible. We trained a world-model-based controller in this environment and deployed the learned controller in the real world on a Pepper robot.
In the simulated tests our policy achieved 94\% success rate in the empty case, 55\% in the 3-human case and 17\% in the 10-human scenario.
In real-life tests, 36 out of 54 (66\%) goals were reached successfully (see Fig.~\ref{fig:irl_test_traj}). In those tests, the number of people varied between 0 and 9, similar to the simulator training conditions. People displayed various behaviors, some ignoring the robot and others blocking its way.
Out of all real tests, only two failures involved a collision of any kind. Once was with a chair leg and the other with a poster stand leg.
All other failures were due to an inability to reach the goal in the allotted time (60s per waypoint). These encouraging results demonstrate the benefit of photorealistic 3D simulation with the proposed NavDreams simulator for transferring camera-based navigation policies to the real world.

\section{Conclusion}
In this work we explored the problem of RL-based navigation among people from only monocular camera. We find that the proposed world-model approaches achieve promising results.
We also find that world-models trained on navigation data are able to capture part of the dynamics of navigation and human movement.
Our multi-task experiment results showed that the world-models encodings are comparatively rich; they can be used to recover not just image data but also geometric understanding (depth) and semantic understanding.
%
%
In our small-scale sim-to-real experiments, we showed that the learned policy performs similarly in the real world as in the simulation, which validates the appropriateness of our proposed simulation environment. 

Future work on the method side should increase the variety of navigation scenarios in order to ensure that overfitting is avoided.
In general, we observe that RL approaches struggle to understand the full scene context, model the future impacts of their actions and use this information to plan ahead. As a result, human-performance is still unmatched on these navigation tasks.
We believe that the world-model prediction quality is key to making the planning task tractable and improving it will be critical in future research.
%



\bibliographystyle{IEEEtran}
\footnotesize
\bibliography{references}




\end{document}